# Contextual Attention Mechanism, SRGAN Based Inpainting System for Eliminating Interruptions from Images

Narayana Darapaneni
Director - AIML
*Great Learning/Northwestern University*
*Illinois, USA*
darapaneni@gmail.com

Deepali Nikam
Student - AIML
*Great Learning*
*Pune, India*
deepalinikam311@gmail.com

Anagha Lomate
Student - AIML
*Great Learning*
*Pune, India*
alomate91@gmail.com

Vaibhav Kherde
Student - AIML
*Great Learning*
*Pune, India*
kherde.vaibhav@gmail.com

Swanand Katdare
Student - AIML
*Great Learning*
*Pune, India*
katdare.swanand@gmail.com

Anwesh Reddy Paduri
Senior Data Scientist- AIML
*Great Learning*
*Pune, India*
anwesh@greatlearning.in

Kameswara Rao
Student - AIML
*Great Learning*
*Pune, India*
raodks@gmail.com

Anima Shukla
Student - AIML
*Great Learning*
*Pune, India*
animashukla0293@gmail.com

*Abstract—* The new alternative is to use deep learning to inpaint any image by utilizing image classification and computer vision techniques. In general, image inpainting is a task of recreating or reconstructing any broken image which could be a photograph or oil/acrylic painting. With the advancement in the field of Artificial Intelligence, this topic has become popular among AI enthusiasts. With our approach, we propose an initial end-to-end pipeline for inpainting images using a complete Machine Learning approach instead of a conventional application-based approach. We first use the YOLO model to automatically identify and localize the object we wish to remove from the image. Using the result obtained from the model we can generate a mask for the same. After this, we provide the masked image and original image to the GAN model which uses the Contextual Attention method to fill in the region. It consists of two generator networks and two discriminator networks and is also called a coarse-to-fine network structure. The two generators use fully convolutional networks while the global discriminator gets hold of the entire image as input while the local discriminator gets the grip of the filled region as input. The contextual Attention mechanism is proposed to effectively borrow the neighbor information from distant spatial locations for reconstructing the missing pixels. The third part of our implementation uses SRGAN to resolve the inpainted image back to its original size. Our work is inspired by the paper Free-Form Image Inpainting with Gated Convolution and Generative Image Inpainting with Contextual Attention.

## I. INTRODUCTION

Photo software code like Photoshop/Corel Paint shop has evolved a heap in the past couple of years. However, to use this software one must purchase the license also have skills beforehand. With growing AI capabilities in Computer Vision, it is currently easier to manipulate snaps consistency with our desires.

With this approach of ours, we are going to strive removing obstructions using machine-learning algorithms from the images and therefore minimizing the manual dependencies on this code.

Image inpainting has invariably been a challenging and ongoing field of exploration for consumers. With the rising demand for taking nice pictures, lots of effort has been endowed in building higher tools for users to require athletically taking photos. However typically, users may take undesirable pictures once their object of interest is impeded by the unrelated.

Most of them use some type of inpainting algorithm to fill in the desired removed part. However even with most cutting-edge inpainting algorithms results on the images are always unpredictable, since its accuracy vastly depends on the background of the image, area of required removed region and clarity, the complexity of the image.

The 3 most used image classification techniques are simple image classifier, that merely tells us what single object or scene is present in the image, object detection that locates and classify multiple objects inside a picture by drawing bounding boxes around them so classifying what's within the box, and at last, semantic segmentation that is the most accurate among 3 where instead of rectangular bounding boxes, the mask is formed to classify each pixel within the image.

Removing undesirable reflections from images captured through glass also can be thought of an obstruction. Though state-of-the-art ways will acquire decent leads to sure things, performance declines considerably once coping with additional general real-world cases. Several imaging conditions are far from best, forcing us to require our photos through reflecting or occluding components. For instance, once taking photos through glass windows, reflections from indoor objects will obstruct the outside scene we want to capture. Similarly, to require pictures of animals within the zoo, we may have to shoot through associate degree enclosure or a fence. Such visual obstructions are usually not possible to avoid simply by changing the camera position or the plane of focus, and state- of-the-art computational approaches are still not sturdy enough to get rid of such obstructions from pictures with ease.

The Places dataset is meant following principles of human visual cognition. The goal is to create a core of visual data that may be used to train artificial systems for high-level visual understanding tasks, like scene context, beholding, action and event prediction, and theory-of-mind inference. The semantic categories of Places square measure outlined by their function: the labels represent the entry-level of an environment. Let us say, the dataset has very different classes of bedrooms, or streets, etc., together does not act the same way, and does not create the same predictions of what will happen next, in a home bedroom, a hotel bedroom, or a nursery.

In total, Places contains more than ten million pictures comprising 400+ unique scene classes. The dataset features 5000 to thirty, 000 training pictures per class, in keeping with real-world frequencies of occurrence. Using convolutional neural networks (CNN), the Places dataset allows learning of deep scene features for numerous scene recognition tasks, to determine new state-of-the-art performances on scene-centric benchmarks.

## II. MATERIALS AND METHODS

### A. Methods and Modelling strategy

Inpainting is a conservation method wherever damaged, deteriorating, or missing components of the artwork are filled- in to furnish an entire image. This method can be applied to each physical and digital art medium. [5]

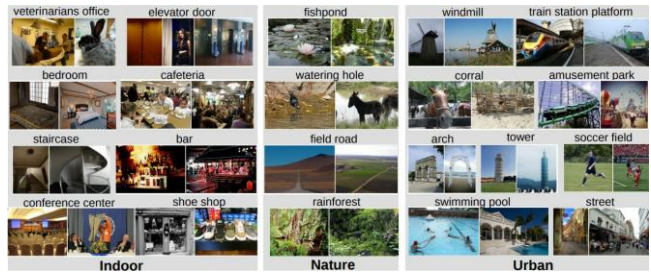

Fig. 1: Places2 dataset sample images

TABLE.1. TOP 5 ACCURACY RESULTS ON PLACES DATASET

|  | Validation Set of Places365 | | Test Set of Places365 | |
| --- | --- | --- | --- | --- |
|  | Top-1 acc. | Top-5 acc. | Top-1 acc. | Top-5 acc. |
| Places365 - AlexNet | 53.17% | 82.89% | 53.31% | 82.75% |
| Places365 - GoogleNet | 53.63% | 83.88% | 53.59% | 84.01% |
| Places365 - VGG | **55.24%** | 84.91% | **55.19%** | 85.01% |
| Places365 - ResNet | 54.74% | **85.08%** | 54.65% | **85.07%** |
|  |  |  |  |  |

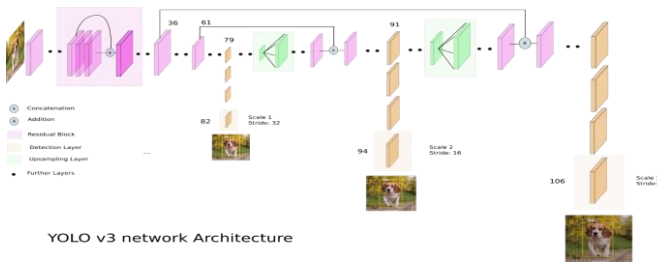

Fig. 2: YOLO V3 network Architecture

Image inpainting is an active area of artificial intelligence (AI) research wherever AI has been able to return up with better inpainting results than most solutions on the market. In this report, we are going to discuss image inpainting using deep learning- based approaches that have shown promising results for this challenging task.

For image inpainting, traditional texture and patch synthesis approaches are particularly appropriate once it needs to borrow textures from the encompassing regions [1]. On the other hand, it is for the most part believed that convolutional neural networks are ineffective in explicitly borrowing or repetition data from distant spatial locations. This led to a spark in the AI community and made it one of the foremost the most topics among researchers.

In this solution, we wished to create whenever we pass the image directly to the object localization algorithm which will create a mask for the known object. after which the masked Images are passed to the Contextual Attention rule which will in-paint the masked region. We conjointly give the output to a different rule to enlarge the whole image.

### B. YOLOv3:

Object detection is a task in computer vision that involves identifying the presence, location, and type of one or additional objects in a given photograph. It is a challenging problem that involves building upon ways for object recognition (e.g., where are they), object localization (e.g., what are their extent), and object classification (e.g., what are they). The "You only Look Once," or YOLO, the family of models are a series of end-to-end deep learning models designed for quick object detection, developed by Joseph

Redmon, et al. and first represented in the 2015 paper titled "You only Look Once: Unified, real-time Object Detection."

The approach involves a single deep convolutional neural network (originally a version of GoogLeNet, later updated and known as DarkNet based on VGG) that splits the input into a grid of cells and every cell directly predicts a bounding box and object classification. A result is an outsized number of candidates bounding boxes that are consolidated into a final prediction by a post-processing step. YOLOv3 is trained on the eighty classes available within the MS COCO dataset [20]. We use the pre-trained dataset to get the masked output. For our solution, we have restricted identification to just one class i.e., "dog".

### C. CONTEXTUAL ATTENTION:

The structure of CNNs cannot effectively model the long-term correlations between the missing regions and information given by distant spatial locations. If you are aware of CNNs, you recognize that the kernel size and the dilation rate control the receptive field at a convolutional layer. Hence, the network must go deeper and deeper to visualize the whole input image. This implies that if we would like to capture the context of an image, we have to rely on deeper layers. However, as we move deeper, we lose the spatial information as deep layers have a smaller spatial size of options. Therefore, there was a requirement to find a way to borrow information from distant spatial locations (i.e., understanding the context of an image) while not going too deep into a network. [8]

Early works attempted to solve the problem using ideas similar to texture synthesis, i.e., by matching and repetition background patches into holes starting from low-resolution to high-resolution or propagating from whole boundaries. These approaches work well particularly in background inpainting tasks and are widely deployed in practical applications. However, as they assume missing patches can be fund somewhere in background regions, they cannot challenge novel image contents for difficult cases wherever inpainting regions involve complex, non-repetitive structures (e.g., faces, objects).

The contextual Attention mechanism is proposed to be effective in borrowing the contextual information from distant spatial locations for reconstructing the missing pixels. The proposed framework consists of two generator networks and two discriminator networks. The two generators follow the fully convolutional networks with expanded convolutions. One generator is for coarse reconstruction and another one is for refinement. This is often known as commonplace coarse-to-fine network structure. The two discriminators also look at the completed images each globally and locally. The global discriminator takes the entire image as input while the local discriminator takes the filled region as input. For the architecture, the proposed framework consists of two generator networks and two discriminator networks. The two generators follow the fully convolutional networks with dilated convolutions.

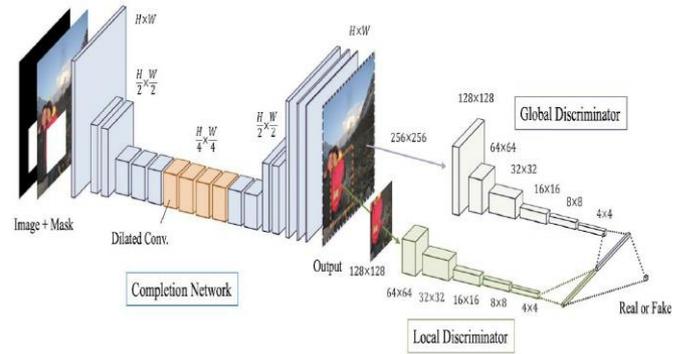

Fig. 3: Contextual Attention Architecture

One generator is for coarse reconstruction and another one is for refinement. This is often known as the standard coarse-to-fine network structure. The two discriminators also look at the finished pictures both globally and locally [12]. The global discriminator takes the whole image as input whereas the local discriminator takes the filled region as input.

We follow the same input and output configurations as in for training and inference, i.e., the generator network takes an image with white pixels filled within the holes and a binary mask indicating the opening regions as input pairs, and outputs the final completed image. We combine the input with a corresponding binary mask to handle holes with variable sizes, shapes, and locations. The input to the network randomly 256*256 image with a rectangle missing region sampled randomly during training, and therefore the trained model can take an image of various sizes with multiple holes in it. A two-stage coarse-to-fine network architecture was introduced and wherever the first network makes an initial coarse prediction, the second network takes the coarse prediction as inputs and predicts refined results. The coarse network is trained with the reconstruction loss explicitly, whereas the refinement network is trained with the reconstruction as well as GAN losses [10]. Intuitively, the refinement network sees a more complete scene than the original image with missing regions; therefore, its encoder [5] can learn better feature representation than the coarse network. In addition, ELUs was used as activation functions instead of ReLU in and clip the output filter values instead of using tanh or sigmoid functions.

### D. LOSS FUNCTION

In the base paper for the loss function, they employ adversarial loss (GAN loss [11]) and L1 loss (for pixel-wise reconstruction accuracy). For the L1 loss, they use a spatially discounted L1 loss in which a weight is assigned to every pixel difference and therefore the weight is based on the distance of a pixel to its nearest known pixel. Gulrajani and his team proposed an improved version of WGAN with a gradient penalty (WGAN-GP). For GAN loss, this WGAN-

GP loss is used rather than the standard adversarial loss. WGAN adversarial loss is additionally based on L1 distance measure, therefore the network is simpler to train and the training process is more stable. This loss attached is then connected to each global and local outputs of the second-stage refinement network to enforce global and local consistency. WGAN-GP loss is well known to outperform existing GAN losses for image generation tasks, and it works well when combined with L1 reconstruction loss as they both use the L1 distance metric. WGAN uses the Earth-Mover distance (a.k.a Wasserstein-1) distance W (Pr, Pg) for comparing the generated and real data distributions.

Its objective function is constructed by applying the Kantorovich-Rubinstein duality:

$$G\ D \in D\ Ex \sim Pr\ [D(x)] - ExPg[D(x)] \quad (1)$$

Where D is the set of 1-Lipschitz functions and Pg is the model, distribution implicitly defined by x = G (z), z is the input to the generator, and WGAN with a gradient penalty term is given by:

$$ExPx(||xD(x)||2 - 1)2 \quad (2)$$

Where x is sampled from the straight line between points sampled from distribution Pg and Pr. The reason is that the gradient of D* at all points x = (1 − t) x + tx on the straight line should point directly towards current sample x, meaning

$$xD*(x) = x - x || x - x || \quad (3)$$

For image inpainting, we predict on masked/cropped regions, thus the gradient penalty should be applied only to pixels which are masked. This can be implemented with the multiplication of gradients and input mask m as follows:

$$ExPx\ (|| \nabla xD(x) \odot (1 - m) || 2 - 1)2 \quad (4)$$

Where the mask value is 0 for missing pixels and 1 for elsewhere. We use a weighted sum of pixel-wise L1 loss and WGAN adversarial losses. Note that in primal space, Wasserstein-1 distance in WGAN is based on L1 ground distance:

$$WPrPg = inf\gamma \in \Pi(Pr\ Pg)E(x,y) \sim \gamma\ [|| x - y ||] \quad (5)$$

where (Pr , Pg) denotes the set of all joint distributions (x, y) whose marginals are respectively Pr and Pg. Instinctively, in the pixel-wise reconstruction the loss function will try to recreate masked pixels to match the ground truth image, while WGANs implicitly learn to match and potentially correct images and also train these generators using adversarial gradients.

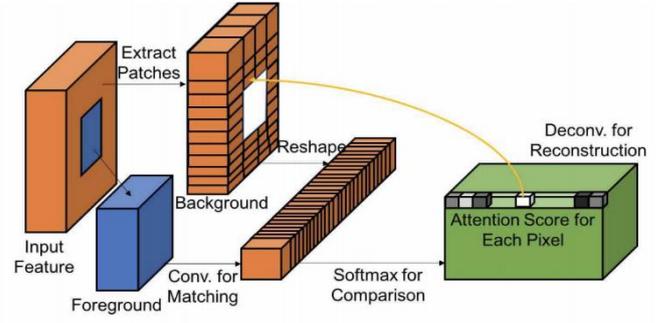

Fig. 4: Working of Contextual Attention layer

*E. SRGAN (Super Resolution GAN)*

The researchers at Twitter proposed SRGAN [15]. The motive of this architecture is to recover finer textures from the image once we upscale it so that its quality cannot be compromised. There are alternative ways methods bilinear Interpolation that can be used to perform this task; however, they suffer from image information loss and smoothing. In this paper, Super Resolution GAN (SRGAN) the authors proposed two architectures, the one without GAN (SRResNet) and one with GAN (SRGAN). It can be concluded that SRGAN has better accuracy and generates images more pleasing to the eyes as compared to SRResNet. The output generated from the contextual Attention is a 256*256 image that we need to convert to the appropriate size. Hence, we use the SRGAN to create a much better output.

III. RESULTS

As discussed earlier, we have three models to look at for making the system robust. There are several object detection models coming from academic papers that portray promising results in terms of mAP and performance. We chose yolov3 pre-trained on the coconut palm dataset containing eighty classes. For image inpainting several state-of-the-art models for object removal. We trained current contextual models i.e., Generative image inpainting with discourse attention.

Our training results have:

- L1 loss leader as 18.9
- L2 loss leader as 5.6
- PSNR (Peak Signal to Noise Ratio) as 16.8
- TV Loss as 28th

We used pre-trained SR-GAN [8] for obtaining the super-resolution.

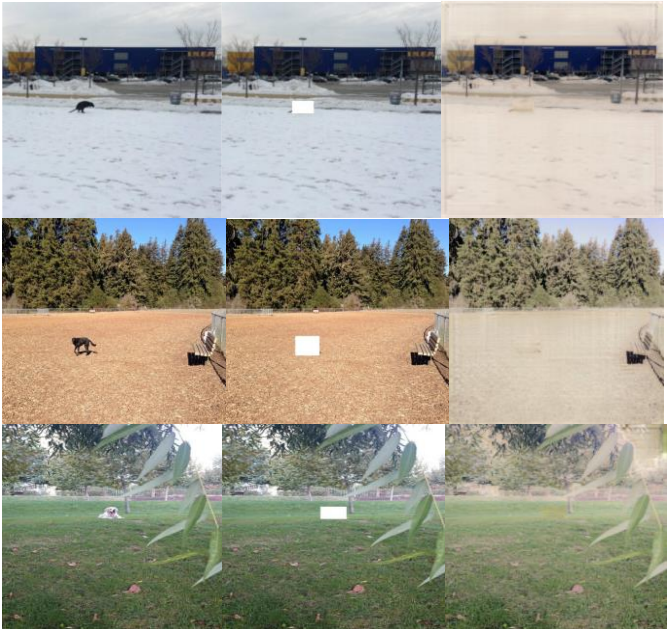

Fig. 5: Pre and Post model result images

## IV. DISCUSSION AND CONCLUSION

The goal of object detection is to determine whether or not there are any instances of objects from given classes (such as humans, cars, bicycles, dogs, or cats) in an image and, if present, to return the spatial location and extent of every object instance.

The abundant usage of mobile devices and social media had increased the demand for visual data analysis. Having said that, there are computational capability limitations of these mobile/wearable devices which make object detection challenging. Moreover, the efficiency is also controlled by need to localize and recognize objects in single image. Computational complexity can also grow with the (possibly large) variety of object classes, and with the immense number of their location in same image.

As the cornerstone of image understanding and computer vision, object detection forms the basis for resolution solving high-level vision tasks like segmentation, scene understanding, object tracking, image captioning, event detection, and activity recognition.

This paper proposes a new method of eliminating unwanted objects from images using a machine learning approach instead of a conventional application-based approach that exhibits good performance in terms of inpainting cropped regions of the images. By using YOLO we can automatically identify the objects and create mask for it. And by using contextual attention, we are effectively borrowing information from distant spatial locations for reconstructing the local missing pixels to get both global semantic structure and finely detailed textures. In alternative words, we are filling in an image and its context is maintained. Finely detailed textures mean that the generated pixels are realistic-looking and as sharp as possible.

Thus, creating an end-to-end pipeline for automatically detecting and removing unwanted irruptions from each image that is able to achieve comparable results to hand-crafted manual target selection.

Because of the limitation in substantial processing power, we could achieve a fair result however, we can achieve better results if we train on more data sets to understand all types of scenarios and situations present in different images using GAN's [6] which can resolve the unsupervised problems very effectively. Also, in our implementation we saw that if the object in the input image is substantially large for the image, then the reconstructed image is distorted on the removed part. Object removal works better on the removal of smaller objects compared to larger objects.

Our future scope includes to generate masks using Freeform selection for image inpainting and to address the limitations mentioned in previous section as well as increase the practical usage.

REFERENCES

[1]. Yu, J., Lin, Z., Yang, J., Shen, X., Lu, X., & Huang, T. S. (2018). Genrative image inpainting with contextual attention. Retrieved from http://arxiv.org/abs/1801.07892
[2]. Goodfellow I, Pouget-Abadie J, Mirza M, Xu B, Warde-Farley D, Ozair S, Courville A, Bengio Y (2014) Generative adversarial nets, In: Advances in neural information processing systems, pp 2672–2680
[3]. Dong J, Yin R, Sun X, Li Q, Yang Y, Qin X (2018) Inpainting of remote sensing sst images with deep convolutional generative adversarial network. IEEE Geosci Remote Sens Lett 16(2):173–177
[4]. Lou S, Fan Q, Chen F, Wang C, Li J, (2018) Preliminary investigation on single remote sensing image inpainting through a modified gan, In: 2018 10th IAPR workshop on pattern recognition in remote sensing (PRRS). IEEE 1–6
[5]. Salem N.M, Mahdi HM, Abbas H (2018) Semantic image inpainting vsing self-learning encoder-decoder and adversarial loss, In: 2018 13th international conference on computer engineering and systems (ICCES), IEEE, pp 103–108
[6]. Liu H, Lu G, Bi X, Yan J, Wang W (2018) Image inpainting based on generative adversarial networks, In: 2018 14th international conference on natural computation, fuzzy systems and knowledge discovery (ICNC-FSKD), IEEE, pp 373–378
[7]. Han X, Wu Z, Huang W, Scott MR, Davis LS (2019) Compatible and diverse fashion image inpainting. arXiv preprint arXiv:1902.01096
[8]. Jiao L, Wu H, Wang H, Bie R (2019) Multi-scale semantic image inpainting with residual learning and GAN. Neurocomputing 331:199–212
[9]. Nazeri K, Ng E, Joseph T, Qureshi F, Ebrahimi M (2019) Edgeconnect: generative image inpainting with adversarial edge learning. arXiv preprint arXiv:1901.00212
[10]. Salem N.M, Mahdi HM, Abbas H (2018) Semantic image inpainting vsing self-learning encoder-decoder and adversarial loss, In: 2018 13th international conference on computer engineering and systems (ICCES), IEEE, pp 103–108
[11]. Li A, Qi J, Zhang R, Ma X, Ramamohanarao K (2019) Generative image inpainting with submanifold alignment. arXiv preprint arXiv:1908.00211
[12]. Li A, Qi J, Zhang R, Kotagiri R (2019) Boosted gan with semantically interpretable information for image inpainting, In: 2019 international joint conference on neural networks (IJCNN), IEEE, pp 1–8
[13]. Armanious K, Mecky Y, Gatidis S, Yang B (2019) Adversarial inpainting of medical image modalities, In: ICASSP 2019–2019 IEEE international conference on acoustics, speech and signal processing (ICASSP) IEEE, pp 3267–3271
[14]. Yeh R A, Chen C, Yian Lim T, Schwing AG, Hasegawa-Johnson M, Do MN (2017) Semantic image inpainting with deep generative

models, In: Proceedings of the IEEE conference on computer vision and pattern recognition, pp 5485–5493
15. Jiao L, Wu H, Wang H, Bie R (2019) Multi-scale semantic image inpainting with residual learning and GAN. Neurocomputing 331:199–212
16. Nazeri K, Ng E, Joseph T, Qureshi F, Ebrahimi M (2019) Edgeconnect: generative image inpainting with adversarial edge learning. arXiv preprint arXiv:1901.00212
17. Li A, Qi J, Zhang R, Ma X, Ramamohanarao K (2019) Generative image inpainting with submanifold alignment. arXiv preprint arXiv:1908.00211
18. Li A, Qi J, Zhang R, Kotagiri R (2019) Boosted gan with semantically interpretable information for image inpainting, In: 2019 international joint conference on neural networks (IJCNN), IEEE, pp 1–8
19. Armanious K, Mecky Y, Gatidis S, Yang B (2019) Adversarial inpainting of medical image modalities, In: ICASSP 2019–2019 IEEE international conference on acoustics, speech and signal processing (ICASSP) IEEE, pp 3267–3271
20. Yeh R A, Chen C, Yian Lim T, Schwing AG, Hasegawa-Johnson M, Do MN (2017) Semantic image inpainting with deep generative models, In: Proceedings of the IEEE conference on computer vision and pattern recognition, pp 5485–5493
21. Lin TY, Maire M, Belongie S, Hays J, Perona P, Ramanan D, Dollár P, Zitnick CL (2014) Microsoft coco: common objects in context. In: European conference on computer vision. Springer, Cham, pp 740–755
22. J. Yu, M. Tan, H. Zhang, D. Tao, and Y. Rui, "Hierarchical Deep Click Feature Prediction for Fine-grained Image Recognition," IEEE Transactions on Pattern Analysis and Machine Intelligence, pp. 1–1, 2019. [Online]. Available: 10.1109/tpami.2019.2932058; https://dx.doi.org/10.1109/tpami.2019.2932058
23. J. Yu, C. Zhu, J. Zhang, Q. Huang, and D. Tao, "Spatial Pyramid-Enhanced NetVLAD With Weighted Triplet Loss for Place Recognition," IEEE Transactions on Neural Networks and Learning Systems, vol. 31, no. 2, pp. 661–674, 2020. [Online]. Available: 10.1109/tnnls.2019.2908982; https://dx.doi.org/10.1109/tnnls.2019.2908982
24. J. Zhang, J. Yu, and D. Tao, "Local deep-feature alignment for unsuper- vised dimension reduction," IEEE Trans Image Process, vol. 27, no. 5, pp. 2420–2432, 2018.
25. J. Yu, D. Tao, M. Wang, and Y. Rui, "Learning to Rank Using User Clicks and Visual Features for Image Retrieval," IEEE Transactions on Cybernetics, vol. 45, no. 4, pp. 767–779, 2015. [Online]. Available: 10.1109/tcyb.2014.2336697; https://dx.doi.org/10.1109/tcyb.2014.2336697
26. J. Yu, X. Yang, F. Gao, and D. Tao, "Deep multimodal distance metric learning using click constraints for image ranking," IEEE Trans Cybern, vol. 47, no. 12, pp. 4014–4024, 2016.
27. C. Hong, J. Yu, J. Wan, D. Tao, and M. Wang, "Multimodal Deep Autoencoder for Human Pose Recovery," IEEE Transactions on Image Processing, vol. 24, no. 12, pp. 5659–5670, 2015. [Online]. Available: 10.1109/tip.2015.2487860; https://dx.doi.org/10.1109/tip.2015.2487860
28. C. Hong, J. Yu, D. Tao, and M. Wang, "Image-based three-dimensional human pose recovery by multiview locality-sensitive sparse retrieval," IEEE Trans Ind Electron, vol. 62, no. 6, pp. 3742–3751, 2014.
29. J. Yu, B. Zhang, Z. Kuang, D. Lin, and J. Fan, "iPrivacy: Image Privacy Protection by Identifying Sensitive Objects via Deep Multi-Task Learning," IEEE Transactions on Information Forensics and Security, vol. 12, no. 5, pp. 1005–1016, 2017. [Online]. Available: 10.1109/tifs.2016.2636090; https://dx.doi.org/10.1109/tifs.2016.2636090
30. C. Hong, J. Yu, J. Zhang, J. X. Lee, and K. H, "Multi-modal face pose estimation with multi-task manifold deep learning," IEEE Trans Ind Inform, vol. 15, no. 7, pp. 3952–3961, 2018.
31. O. Elharrouss, A. Abbad, D. Moujahid, J. Riffi, and H. Tairi, "A block-based background model for moving object detection," Electron Lett Comput Vis Image Anal, vol. 15, no. 3, pp. 17–31, 2016.
32. A. Abbad, O. Elharrouss, K. Abbad, and H. Tairi, "Application of MEEMD in post-processing of dimensionality reduction methods for face recognition," IET Biometrics, vol. 8, no. 1, pp. 59–68, 2019. [Online]. Available: 10.1049/iet-bmt.2018.5033; https://dx.doi.org/10.1049/iet-bmt.2018.5033
33. D. Moujahid, O. Elharrouss, and H. Tairi, "Visual object tracking via the local soft cosine similarity," Pattern Recognition Letters, vol. 110, pp. 79–85, 2018. [Online]. Available: 10.1016/j.patrec.2018.03.026; https://dx.doi.org/10.1016/j.patrec.2018.03.026
34. S. M. Muddala, R. Olsson, and M. Sjöström, "Spatio-temporal consistent depth-image-based rendering using layered depth image and inpainting," EURASIP Journal on Image and Video Processing, vol. 2016, no. 1, pp. 9–9, 2016. [Online]. Available: 10.1186/s13640-016-0109-6; https://dx.doi.org/10.1186/s13640-016-0109-6
35. M. Isogawa, D. Mikami, D. Iwai, H. Kimata, and K. Sato, "Mask Optimization for Image Inpainting," IEEE Access, vol. 6, pp. 69 728–69 741, 2018. [Online]. Available: 10.1109/access.2018.2877401; https://dx.doi.org/10.1109/access.2018.2877401T.
36. Ružić and A. Pižurica, "Context-aware patch-based image inpainting using markov random field modeling," IEEE Trans Image Process, vol. 24, no. 1, pp. 444–456, 2014.
37. K. H. Jin and J. C. Ye, "Annihilating filter-based low-rank hankel matrix approach for image inpainting," IEEE Trans Image Process, vol. 24, no. 11, pp. 3498–3511, 2015.
38. N. Kawai, T. Sato, and N. Yokoya, "Diminished Reality Based on Image Inpainting Considering Background Geometry," IEEE Transactions on Visualization and Computer Graphics, vol. 22, no. 3, pp. 1236–1247, 2016. [Online]. Available: 10.1109/tvcg.2015.2462368; https://dx.doi.org/10.1109/tvcg.2015.2462368
39. Q. Guo, S. Gao, X. Zhang, Y. Yin, and C. Zhang, "Patch-based image inpainting via two-stage low rank approximation," IEEE Trans Vis Comput Gr, vol. 24, no. 6, pp. 2023–2036, 2017.
40. H. Lu, Q. Liu, M. Zhang, Y. Wang, and X. Deng, "Gradient-based low rank method and its application in image inpainting," Multimed Tools Appl, vol. 77, no. 5, pp. 5969–5993, 2018.
41. H. Xue, S. Zhang, and D. Cai, "Depth Image Inpainting: Improving Low Rank Matrix Completion With Low Gradient Regularization," IEEE Transactions on Image Processing, vol. 26, no. 9, pp. 4311–4320, 2017. [Online]. Available: 10.1109/tip.2017.2718183; https://dx.doi.org/10.1109/tip.2017.2718183
42. J. Liu, S. Yang, Y. Fang, and Z. Guo, "Structure-guided image inpainting using homography transformation," IEEE Trans Multimed, vol. 20, no. 12, pp. 3252–3265, 2018.
43. D. Ding, S. Ram, and J. J. Rodriguez, "Image Inpainting Using Nonlocal Texture Matching and Nonlinear Filtering," IEEE Transactions on Image Processing, vol. 28, no. 4, pp. 1705–1719, 2019. [Online]. Available: 10.1109/tip.2018.2880681; https://dx.doi.org/10.1109/tip.2018.2880681
44. J. Duan, Z. Pan, B. Zhang, W. Liu, and X. C. Tai, "Fast algorithm for color texture image inpainting using the non-local CTV model," J Glob Optim, vol. 62, no. 4, pp. 853–876, 2015.
45. Q. Fan and L. Zhang, "A novel patch matching algorithm for exemplar-based image inpainting," Multimed Tools Appl, vol. 77, no. 9, pp. 10 807–10 821, 2018.
46. W. Jiang, "Rate-distortion optimized image compression based on image inpainting," Multimed Tools Appl, vol. 75, no. 2, pp. 919–933, 2016